%% file: Template.tex
\documentclass{article}
\usepackage{spconf,amsmath,graphicx}
\usepackage{amssymb}
\usepackage{booktabs}

\title{Zero-shot Personalization of objects via textual inversion}
\name{Aniket Roy$^{1}$, Maitreya Suin$^{2}$, Rama Chellappa$^{1}$ }
\address{$^{1}$Johns Hopkins University, $^{2}$Samsung AI Center Toronto}

\begin{document}
%
\maketitle
\begin{abstract}

Recent advances in text-to-image diffusion models have substantially improved the quality of image customization, enabling the synthesis of highly realistic images. Despite this progress, achieving fast and efficient personalization remains a key challenge, particularly for real-world applications. Existing approaches primarily accelerate customization for human subjects by injecting identity-specific embeddings into diffusion models, but these strategies do not generalize well to arbitrary object categories, limiting their applicability.
To address this limitation, we propose a novel framework that employs a learned network to predict object-specific textual inversion embeddings, which are subsequently integrated into the UNet timesteps of a diffusion model for text-conditional customization. This design enables rapid, zero-shot personalization of a wide range of objects in a single forward pass, offering both flexibility and scalability. Extensive experiments across multiple tasks and settings demonstrate the effectiveness of our approach, highlighting its potential to support fast, versatile, and inclusive image customization. To the best of our knowledge, this work represents the first attempt to achieve such general-purpose, training-free personalization within diffusion models, paving the way for future research in personalized image generation.
\end{abstract}

\input{Section/1_Introduction}
\input{Section/2_Related_work}
\input{Section/3_Method}
\input{Section/4_Experiments}

\input{Section/5_Conclusion}

\input{Section/6_Supple}
\bibliographystyle{IEEEbib}
\bibliography{refs}

\end{document}

%% file: Section/1_Introduction.tex
\section{Introduction}

\begin{figure}
    \centering
    \includegraphics[width=0.45\textwidth]{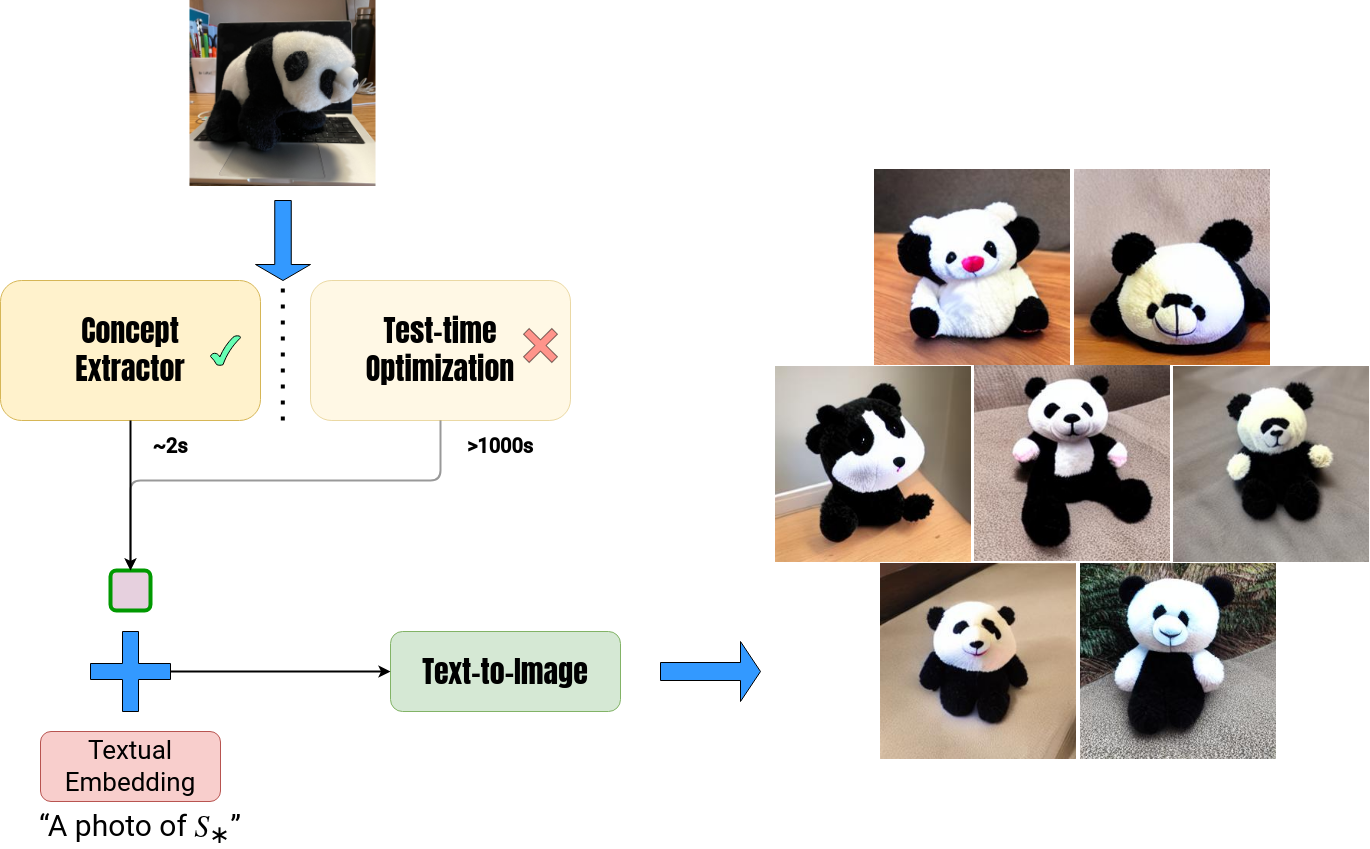}
    \vspace{-0.2cm}
    \caption{\small{Using only a single input image and a concept extraction network, our method is able to personalize a text-to-image diffusion model much faster than test-time optimzation-based approaches, while maintaining the subject’s uniqueness and details.}}
    \label{fig:teaser}
    \vspace{-0.5cm}
\end{figure}

Diffusion models have significantly advanced the field of image generation and editing for their exceptional image quality and variety~\cite{ruiz2022dreambooth,kumari2022multi,roy2023diffnat,roy2022diffalign, roy2025cap2aug, roy2024bri3l, roy2022felmi, roy2026learning}. Recent trends show a growing interest in identity-preserving customization, such as image animation and virtual try-on, due to their wide range of practical applications. Yet, generating varied and high-quality images that maintain identity, object integrity, and textual relevance from only a few examples of a specific object remains a complex challenge. Popular methods like Dreambooth~\cite{ruiz2022dreambooth} and Custom diffusion~\cite{kumari2022multi} have addressed this by either fully or partially fine-tuning the diffusion model. However, these methods come with notable limitations, including (1) the requirement of substantial time (10-15 minutes) for fine-tuning large-scale models, which are typically several gigabytes in size, and (2) a sensitivity to hyperparameters, making them susceptible to overfitting. Such drawbacks limit the applicability of these techniques in settings where resources are limited.


Zero-shot customization has emerged as a significant area of interest, focusing on the adaptation of individual images to match specific text prompts through a single forward pass. This approach is particularly compelling for editing human images, spurred by the expanding applications in augmented and virtual reality, leading to a concentration of research efforts on human subject customization. A noteworthy example is Photomaker~\cite{li2023photomaker}, which introduced the concept of using a stacked image ID as a unique marker for fine-tuning diffusion models, enabling them to process an image and a text prompt in one pass. This was facilitated by a data collection pipeline oriented around ID conservation, specifically designed for humans. However, the challenge of assigning identifiers to generic objects remains, primarily because (1) detailed datasets covering a broad spectrum of objects are scarce, and (2) objects lack a unified identity domain. For instance, while it may be feasible to train and apply a human ID-network across different human subjects, this method is not transferable to generic objects; a model trained on cats and dogs would struggle to identify an airplane with similar precision. Our research addresses the intricate task of customizing any object in a zero-shot setting, a topic that remains largely unexplored in existing literature. Our objective is to devise a method for learning object identifiers during training, enabling the zero-shot customization of test images based on textual prompts through a single forward operation.



We introduce a dual-phase strategy for object customization inspired by recent advances in textual inversion~\cite{gal2022image}. Textual inversion learns a dedicated textual token from a set of training images, which can then be used within a frozen text-to-image diffusion model to enable prompt-guided editing. Rather than aiming to improve the identity accuracy of Textual Inversion, our objective is to remove its reliance on expensive test-time optimization. To this end, we propose a unified formulation that connects Textual Inversion~\cite{gal2022image} and Photomaker~\cite{li2023photomaker} representations, where text tokens function as a generic identity space, enabling fully zero-shot prediction without any per-instance optimization. Building on the idea introduced in Photomaker for human subjects—where a person’s ID serves as a unique marker—we train an MLP to associate images and text prompts with their corresponding inversion tokens, allowing these tokens to act as identifiers for a broad range of objects.

To ensure faithful object portrayal, we fine-tune the cross-attention components of the diffusion model using our dataset. This refinement is primarily aimed at adapting the network to operate on the tokens predicted by the MLP, rather than on the original optimization-based inversion tokens. During zero-shot inference, we first estimate the corresponding textual inversion token for a given image. The predicted token, together with the user text prompt, is then fed into the modified text-to-image diffusion model to generate the customized output. This approach broadens the scope of image personalization by treating textual inversion tokens as universal identifiers, while selectively adjusting model components to ensure authentic and consistent representation of objects in the synthesized images.

In summary, our major contributions are:
\begin{itemize}
\itemsep0em
\item We address the problem of personalization of images of any objects in zero-shot setting in a single forward pass. Compared to existing methods, this setup is more challenging and practically relevant, and to our knowledge has not been explicitly addressed in prior work.
\item We propose a two-stage training method, which involves (1) learning textual inversion mapping for corresponding object identifier, and (2) efficient finetuning of text-to-image diffusion model efficient zero-shot evaluation.
\item We evaluate our approach on text-based image editing for several benchmarks, and shown the efficacy of our approach over SOTA.
\end{itemize}

%% file: Section/2_Related_work.tex
\section{Related Work}

Recent works focused on personalizing text-to-image diffusion models~\cite{roy2023diffnat, roy2025duolora, roy2025multlfg} using limited examples, a challenging problem that requires simultaneously preserving subject identity while maintaining visual realism. Early approaches such as Textual Inversion \cite{gal2022image} and DreamBooth \cite{ruiz2022dreambooth} achieve personalization through test-time optimization, either by introducing new tokens in the text embedding space or by embedding subjects directly into the model’s output space, with Custom Diffusion \cite{kumari2022multi} extending this paradigm to multi-concept composition. While effective, these methods rely on per-concept fine-tuning, making them computationally expensive and limiting scalability. To address this, recent methods explore training-free or single-pass personalization, enabling rapid customization from one image without optimization. Examples include PhotoMaker \cite{li2023photomaker}, which uses stacked ID embeddings for realistic human synthesis, Taming Encoder \cite{jia2023taming} for zero-shot customization, and ELITE \cite{wei2023elite}, which encodes visual concepts into textual embeddings for efficient personalized generation.

%% file: Section/3_Method.tex
\section{Method}



\subsection{Problem statement}


We tackle the issue of personalization with limited samples, specifically aiming to create authentic renditions of objects based on just a few examples while ensuring the output aligns with a given text prompt. Unlike approaches such as Dreambooth \cite{ruiz2023dreambooth} that rely on fine-tuning, our technique operates without any tuning, editing images to match text descriptions in a zero-shot manner through a single forward pass. Even Textual Inversion requires test-time optimization to obtain the text embedding, whereas our approach does not. The main challenge lies in maintaining the integrity of the object and its coherence with the text. Our approach extends beyond the commonly explored domain of human subjects, like in Photomaker \cite{li2023photomaker}, to include a wider array of objects, making it a more comprehensive and demanding task. Although identifying humans in images has been effectively achieved and demonstrated to be adaptable, identifying generic objects poses a greater difficulty. To our knowledge, this work addresses the zero-shot customization of generic objects in a single forward pass, a setting not explicitly explored in prior work. 
We adopt textual inversion tokens as object identifiers and learn an MLP that predicts them in one shot from an image. This predicted token, combined with the text prompt, directly conditions the diffusion model to produce personalized variations in a single pass.


\subsection{Training}

In a zero-shot context, the objects present in the training and testing datasets are distinct, prompting us to utilize a larger and more diverse training set. This diversity is crucial for ensuring the model's robust generalization to novel objects it encounters during testing. To address this demanding task, our approach is structured around a two-phase training methodology. Initially, (1) we focus on training a mapper that establishes a link between an image and its respective textual inversion token, effectively using these tokens as unique identifiers for objects. Subsequently, (2) we refine the diffusion model by fine-tuning its cross-attention blocks in conjunction with the textual inversion tokens, aiming to enhance the accuracy with which subjects are represented. The specifics of this process are elaborated on in the following sections.

\subsection{Learning object identifiers using textual inversion}
\textbf{Preliminaries: Diffusion models.}
Diffusion models~\cite{ho2020denoising, ramesh2022hierarchical, rombach2022high} learn a data distribution by progressively corrupting an image with Gaussian noise and training a denoising network to reverse this process. Given a clean image $x$, noise is added to obtain a noisy sample $x_t$ at timestep $t$, and a denoising model $\epsilon_\theta(x_t,t)$ is trained to predict the injected noise by minimizing
\begin{equation}
\mathcal{L}_{DM}
= \mathbb{E}_{x,\epsilon \sim \mathcal{N}(0,1),t}
\left[
\left\lVert
\epsilon - \epsilon_\theta(x_t,t)
\right\rVert_2^2
\right].
\end{equation}
To improve computational efficiency, latent diffusion models (LDMs) perform the diffusion process in a lower-dimensional latent space. An encoder $G(\cdot)$ maps the input image to a latent representation $z_t = G(x_t)$, and the denoising objective becomes
\begin{equation}
\mathcal{L}_{LDM}
=
\mathbb{E}_{G(x),\epsilon \sim \mathcal{N}(0,1),t}
\left[
\left\lVert
\epsilon - \epsilon_\theta(z_t,t)
\right\rVert_2^2
\right].
\end{equation}
For text-conditioned image generation, a text encoder $\tau_\theta(y)$ maps a textual prompt $y$ into an embedding that is injected into the UNet denoiser via cross-attention layers. The conditional latent diffusion objective is then given by
\begin{equation}
\mathcal{L}_{LDM}
=
\mathbb{E}_{G(x),y,\epsilon \sim \mathcal{N}(0,1),t}
\left[
\left\lVert
\epsilon - \epsilon_\theta(z_t,t,\tau_\theta(y))
\right\rVert_2^2
\right].
\end{equation}

\begin{figure}[!t]
    \centering
    \includegraphics[width=0.5\textwidth]{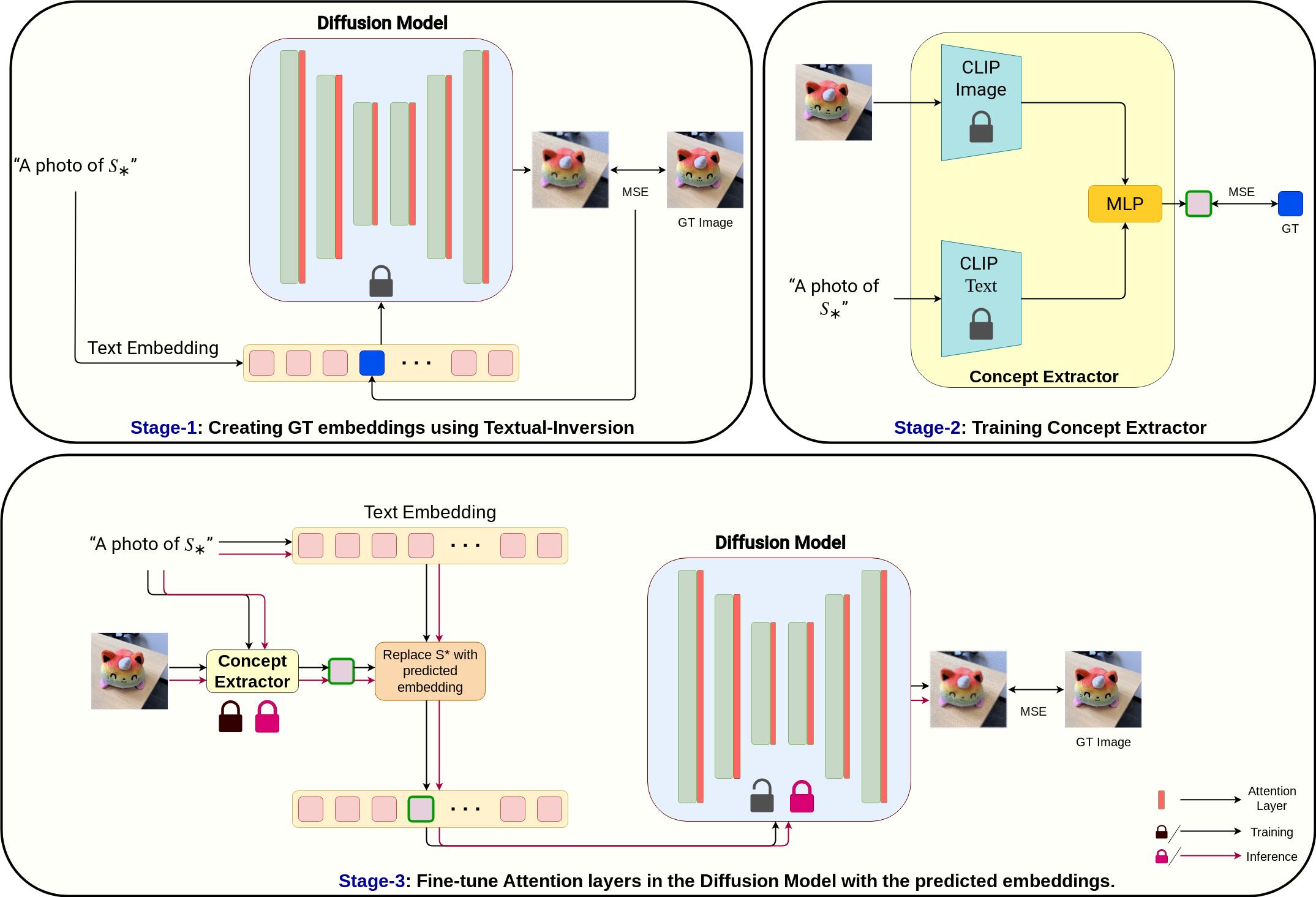}
    \vspace{-0.4cm}
    \caption{An overview of our proposed approach. We first obtain the `ground-truth' textual-embedding of the concepts present in the training set using test-time optimization based-methods like Textual-Inversion. Next, we train our Concept-Extraction network, to produce that embedding given a single image and a text-template. Once it is trained, we further fine-tune the cross-attention layers of the diffusion model using the modified textual-embeddings obtained from the frozen Concept-Extraction network. During inference, the Concept-Extraction network along with the fine-tuned diffusion model can be used to generate variations of the subject present in a given image, without requiring expensive optimization steps.}
    \label{fig:architecture}
    \vspace{-0.5cm}
\end{figure}

\noindent \textbf{Textual inversion.} 
Textual inversion attempts to learn a new concept in form of a placeholder string (``*''), which lead to the reconstruction of images of that particular identity from small set of images. In a way, textual inversion captures the unique identifier a particular object, which motivates us to use textual inversion embedding as an unique identifier of the object. The unique token $v_{*}$ is learned through direct optimization by minimizing the LDM loss over images sampled over small set as follows:
\begin{equation}
    v_{*} = \text{argmin}_{v} \mathbb{E}_{G(x),y,\epsilon \sim \mathcal{N}(0,1),t} [ || \epsilon - \epsilon_{\theta}(z_t, t, \tau_{\theta}(y))||_{2}^{2} ] 
\end{equation}

\begin{figure}
    \centering
    \includegraphics[scale=0.45]{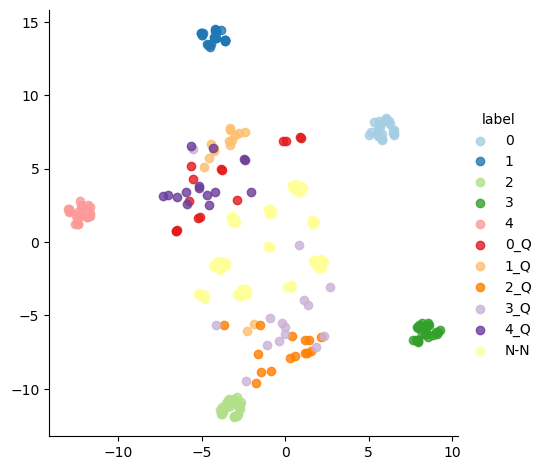}
    \vspace{-0.3cm}
    \caption{\small{tSNE plot for textual inversion embeddings. We observe that the textual inversion embeddings of images of same identity are clustered.}} 
    \label{fig:tsne_ti}
    \vspace{-0.5cm}
\end{figure}

\noindent \textbf{Learning textual inversion mapping.} 
We leverage textual inversion using its standard optimization procedure to learn a text embedding that serves as an identifier for each object. Specifically, for every image in the training set, we compute a corresponding TI embedding through conventional per-instance optimization, thereby constructing a paired dataset of images and their learned textual tokens. These embeddings capture object-specific characteristics, as illustrated by the t-SNE visualization in Fig.~\ref{fig:tsne_ti}. Building on this dataset, we develop a lightweight 3-layer MLP ($f_{\theta}$) to learn a direct mapping from image content to this TI-based identity representation, formulated as $v_{*} = f_{\theta}(I, T)$, where $I$ and $T$ denote the image and its associated text prompt. To increase diversity and augment the training instances, we employ neutral contextual templates such as ``A photo of $v_{*}$,'' ``A close-up of $v_{*}$,'' and ``A random crop of $v_{*}$,'' sampled from the prompt set of \cite{gal2022image}. In practice, we concatenate CLIP image and text embeddings and feed the unified representation to the MLP, which effectively predicts the corresponding textual inversion token.

Our approach aims to develop a concept extractor that requires a single forward pass. Given an image and a text template, it's tasked with generating the corresponding concept in textual-embedding space without requiring test-time optimization. This objective is notably more challenging and ambiguous compared to methods like \cite{gal2022image}, which rely on resource-intensive iterative optimization during testing. We've observed that our concept extractor network faces difficulties when trained to produce the target embedding from scratch, sometimes resulting in divergent training processes and nonsensical textual-embedding outputs.

To enhance the stability of our training regimen, we've adopted the concept of residual learning. We initialize the output using the embedding of a commonly chosen word, such as ``object'' and then train the concept extractor network to learn the `delta' difference in the embedding space, relative to this initialization point. Empirically, this approach significantly stabilizes the training process and yields superior results.
\vspace{-0.5cm}
\subsection{Finetune the Cross-attention blocks of diffusion model using stacked textual inversion.}

In the second phase of our approach, we fine-tune the diffusion model using the training samples, motivated by two key considerations. First, object identifiers learned through textual inversion alone often exhibit lower subject fidelity compared to fine-tuning–based methods such as DreamBooth and Custom Diffusion; however, since test-time fine-tuning is infeasible in our setting, we incorporate diffusion model fine-tuning during training. Second, fine-tuning improves image–text consistency by better aligning image representations with textual inversion embeddings in a shared embedding space.

This raises the question of whether to fine-tune the entire model or only selected components. Prior work by Kumari et al. \cite{kumari2022multi} shows that text-driven edits primarily operate through the cross-attention layers of diffusion models. Accordingly, we restrict fine-tuning to these cross-attention modules, which both simplifies training and reduces the risk of overfitting.

\vspace{-0.5cm}
\subsection{Zero-shot inference in a single forward pass.}



Our objective is to enable zero-shot personalization from a single image or a small set of images using a single forward pass. To this end, we employ a trained MLP ($f_{\theta}$) to predict the corresponding textual inversion from an input image (or image–text pair), together with a fine-tuned diffusion model ($\epsilon_{\theta}$). Given a test image $I_{\text{test}}$, the MLP first produces its textual inversion embedding $v_{\text{test},*}$, which is then combined with the prompt embedding $p$. This combined representation is subsequently propagated through the diffusion model across all timesteps to generate the personalized output images in a single forward pass.
\begin{equation}
    v_{test, *} = f_{\theta} (I_{test}), \quad
    I_{edit} = \epsilon_{\theta} (v_{test, *}, p) 
\end{equation}

%% file: Section/4_Experiments.tex
\section{Experiments}
\vspace{-0.5cm}

\begin{figure}
    \centering
    \includegraphics[scale=0.25]{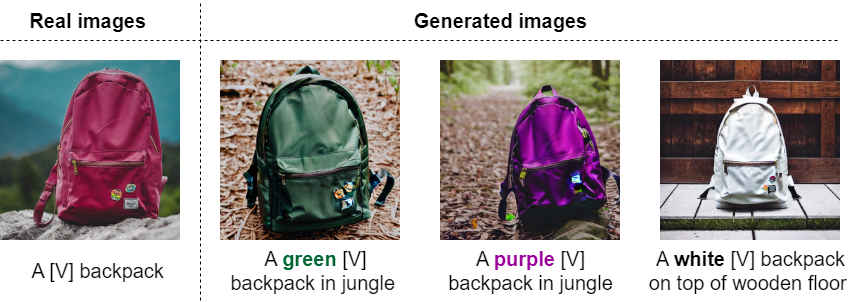}
    \vspace{-0.3cm}
    \caption{\small{Qualitative results on color modification.}} 
    \label{fig:zsp_color_mod}
    \vspace{-0.5cm}
\end{figure}

\begin{figure}
    \centering
    \includegraphics[scale=0.2]{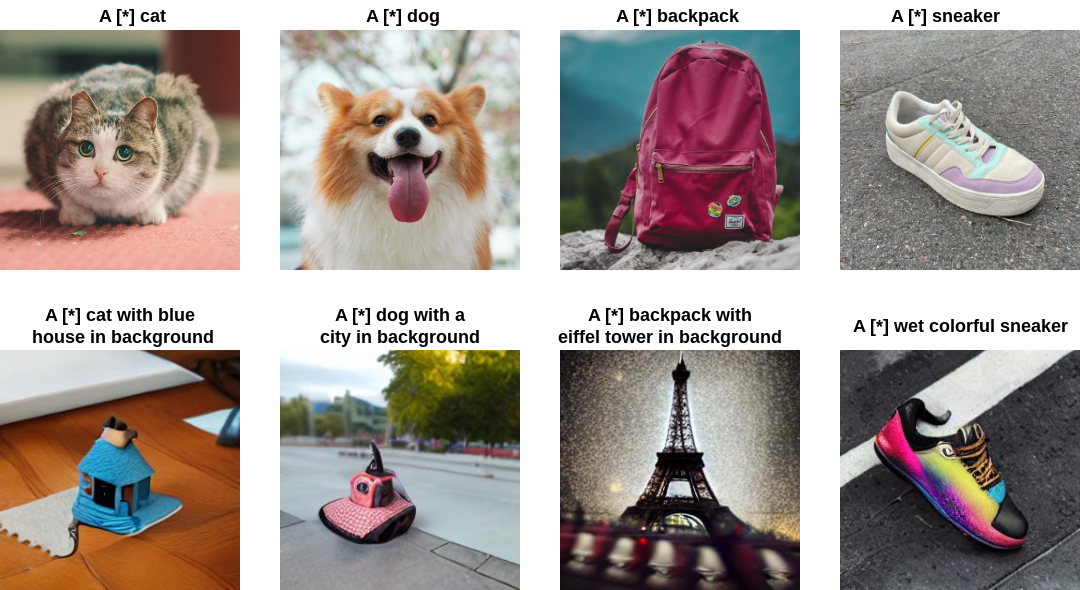}
    \vspace{-0.3cm}
    \caption{\small{Failure cases}} 
    \label{fig:failure_cases}
    \vspace{-0.5cm}
\end{figure}

\noindent\textbf{Experimental Setup.} We evaluate our method in a zero-shot setting with disjoint training and test objects using the Custom101 dataset \cite{kumari2022multi}. Specifically, we train on 71 object categories and test on the remaining 30 unseen categories, with split details provided in the supplementary material. To further assess generalization, we also evaluate the trained model on the DreamBooth dataset in a zero-shot manner.
\newline\textbf{Evaluation dataset.}
We conduct experiments on two datasets: Custom101 and DreamBooth. Custom101 contains diverse object categories with 40 images and 20 prompts per concept, while DreamBooth includes 30 subjects evaluated using 25 prompts comprising re-contextualization and property-modification settings. 
\newline\textbf{Evaluation Metric.} Following Dreambooth~\cite{ruiz2022dreambooth}, we use DINO and CLIP-I for subject fidelity and CLIP-T for prompt fidelity. DINO, which is the average pairwise cosine similarity between the ViT-S/16 DINO embeddings~\cite{caron2021emerging} of the generated and real images. CLIP-I, i.e., the average pairwise cosine similarity between CLIP~\cite{radford2015unsupervised} embeddings of the generated and real images. 
To measure the prompt fidelity, we use CLIP-T, which is the average cosine similarity between prompt and image CLIP embeddings.
\newline\textbf{Implementation details.}
We adopt Latent Diffusion Model (LDM) \cite{rombach2022high} for textual inversion. To learn the object identifier, we extract and concatenate CLIP image and text features, followed by a 3-layer MLP with batch normalization and ReLU activations. The model is trained using AdamW (batch size 32, learning rate $10^{-3}$), after which we fine-tune only the cross-attention layers of the diffusion model for 20 epochs with a reduced learning rate of $10^{-5}$ to mitigate overfitting. During inference, images are generated using 50-step DDIM sampling with a classifier-free guidance scale of 10.
\newline\textbf{Baselines.}
We benchmark our personalization approach against a diverse set of baselines, including DreamBooth \cite{ruiz2022dreambooth}, Custom Diffusion \cite{kumari2022multi}, Textual Inversion \cite{gal2022image}, Re-Imagen \cite{chen2022re}, ELITE \cite{wei2023elite}, BLIP-Diffusion \cite{li2024blip}, and Subject-Diffusion \cite{ma2023subject}. These methods differ fundamentally in their assumptions, with DreamBooth, Custom Diffusion, and Textual Inversion relying on fine-tuning and typically achieving higher subject fidelity, while Re-Imagen and ELITE operate in a zero-shot setting and are more directly comparable to our approach. BLIP-Diffusion and Subject-Diffusion also enable zero-shot personalization but leverage additional external models (e.g., BLIP2, Grounded DINO, SAM), yielding stronger performance in some cases. We also show results on zero-shot color modification in Fig.~\ref{fig:zsp_color_mod}.
\begin{figure}
    \centering
    \includegraphics[scale=0.25]{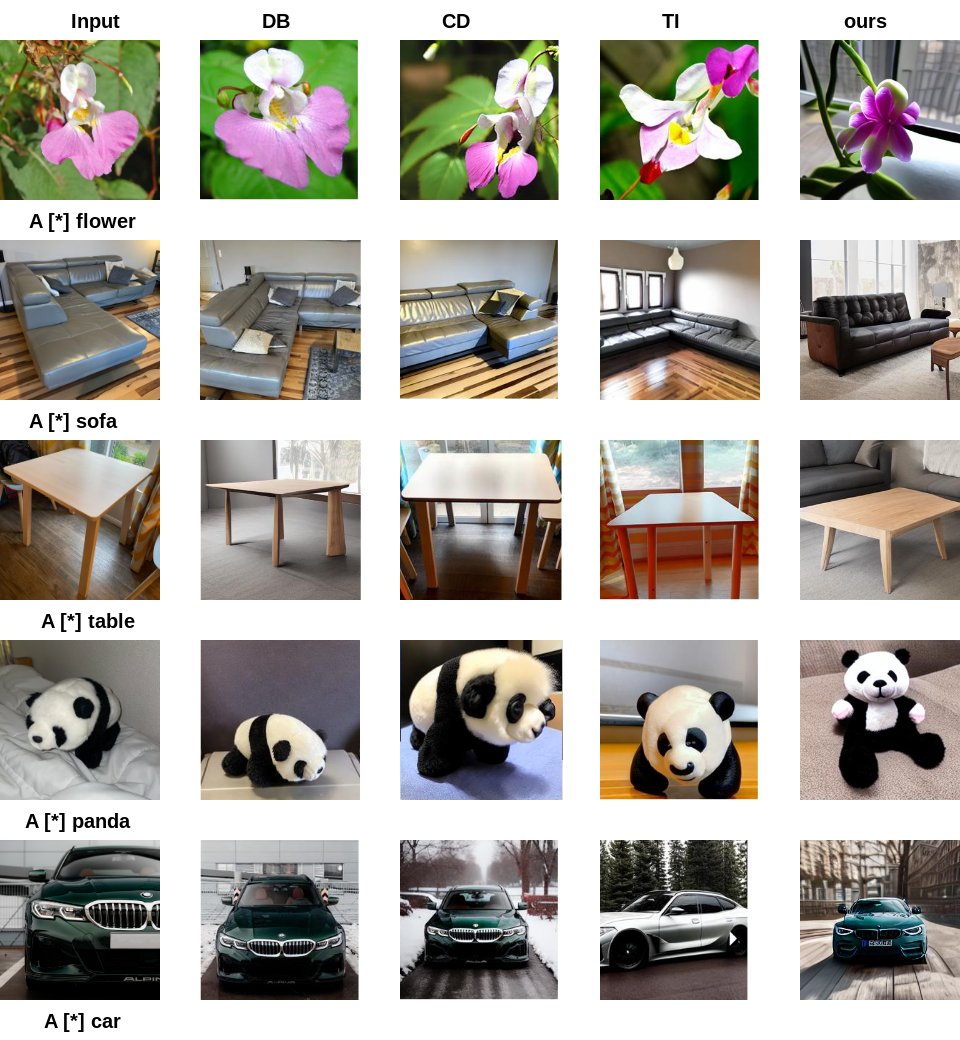}
    \vspace{-0.3cm}
    \caption{\small{Comparison with existing methods (Dreambooth (DB), Custom diffusion (CD), Textual inversion (TI)) on Custom101 dataset.}} 
    \label{fig:zsp_compare_custom101}
    \vspace{-0.5cm}
\end{figure}
\begin{figure}[!t]
    \centering
    \includegraphics[scale=0.31]{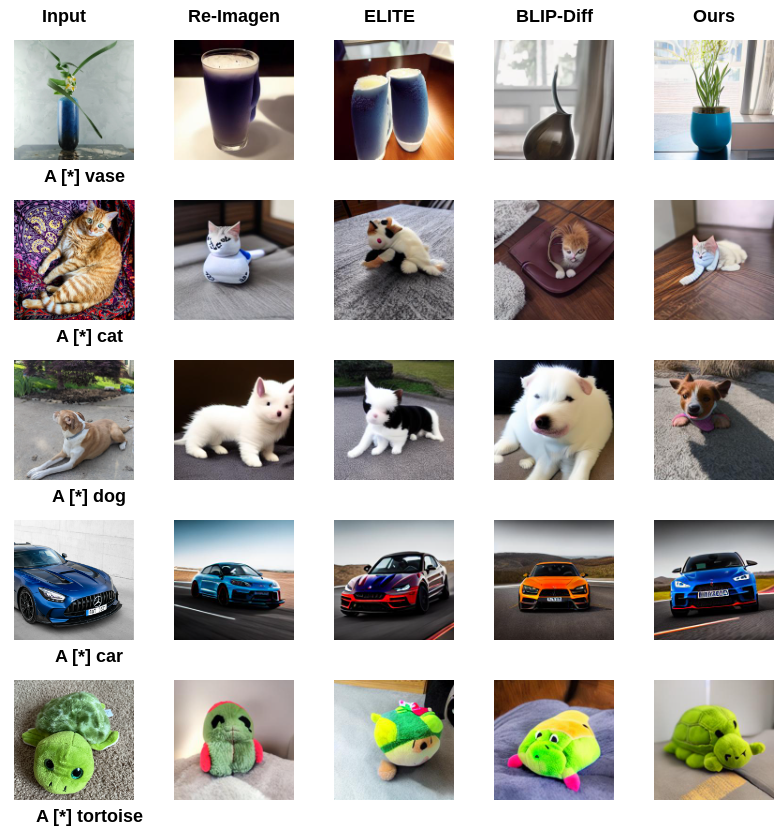}
    \vspace{-0.3cm}
    \caption{\small{Qualitative comparison with existing methods on Custom101 dataset.}} 
    \label{fig:zsp_exp_1}
    \vspace{-0.5cm}
\end{figure}
\newline \textbf{Timing analysis.} Our zero-shot method is significantly faster than the finetuning methods as shown in Table.~\ref{table:zsp_compare_custom101}, i.e., 1200X than textual inversion.
\newline\textbf{Human evaluation.} We additionally conduct human evaluations on Amazon Mechanical Turk to assess subject and text fidelity. Results show that our training-free method preserves subject fidelity in 60\% of cases over 1,500 random trials, with further details provided in the supplementary material.
\newline\textbf{Failure cases.}
Fig.~\ref{fig:failure_cases} shows failure cases where zero-shot conditioning cannot fully preserve subject/prompt fidelity. E.g., given the prompt ``A [*] cat with blue house in background'' the model generates only a blue house, losing the intended cat identity.

\begin{table}[!h]
\small
\centering
\vspace{-0.3cm}
\caption{\small{Comparison of few-shot finetuning task on DB dataset}}
\scalebox{0.65}{
\begin{tabular}{lcccccccc}
\toprule
\hline
\textbf{Method} & {\textbf{Type}} & {\textbf{Ext. Model}} & \textbf{{Test set}} & \textbf{{DINO}} & \textbf{{CLIP-I}} & \textbf{{CLIP-T}} \\
\hline
{Real img.} & {-} & {-} & {-} & {0.774} & {0.885} & {-} \\
{Text. Inv.~\cite{gal2022image}} & {Finetune} & {-} & {DB} & {0.569} & {0.780} & {0.255}\\
{DB~\cite{ruiz2022dreambooth}} & {Finetune} & {-} & {DB} & {0.668} & {0.803} & {0.305} \\
{CD~\cite{kumari2022multi}} & {Finetune} & {-} & {DB} & {0.643} & {0.790} & {0.305} \\
{DiffNat~\cite{roy2023diffnat}} & {Finetune} & {-} & {DB} & {0.68} & {0.84} & {0.34} \\
\hline
{Re-Imagen~\cite{chen2022re}} & {Zero-shot} & {-} & {DB} & {0.600} & {0.740} & {0.270}\\
{ELITE~\cite{wei2023elite}} & {Zero-shot} & {-} & {DB} & {0.621} & {0.771} & {0.293}\\
{BLIP-Diff.~\cite{li2024blip}} & {Zero-shot} & {-} & {DB} & {0.594} & {0.779} & {0.300} \\
{Sub-Diff.~\cite{ma2023subject}} & {Zero-shot} & {BLIP2, GroundedDINO, SAM} & {DB} & {0.711} & {0.787} & {0.293} \\
{Ours} & {Zero-shot} & {-} & {DB} & {0.670} & {0.770} & {0.280} \\
\hline
\bottomrule
\end{tabular}}
\label{table:zsp_compare_db_dataset}
\end{table}
\vspace{-0.5cm}
\begin{table}[!h]
\small
\centering
\vspace{-0.5cm}
\caption{\small{Comparison of few-shot finetuning task on Custom101}}
\scalebox{0.85}{
\begin{tabular}{lccccc}
\toprule
\hline
\textbf{Method} & \textbf{{Type}} & \textbf{{Text-alignment}} & \textbf{{Img-alignment}} & \textbf{{speed(s)}} \\
\hline
{Text. Inv.~\cite{gal2022image}} & {Finetune} & {0.612} & {0.752} & {2400}\\
{DB~\cite{ruiz2022dreambooth}} & {Finetune} & {0.752} & {0.752} & {1284}\\
{CD~\cite{kumari2022multi}} & {Finetune} & {0.769} & {0.744} & {400}\\
{Ours} & {Zero-shot} & {0.590} & {0.625} & {2} \\
\hline
\bottomrule
\end{tabular}}
\label{table:zsp_compare_custom101}
\end{table}
\vspace{-0.5cm}
\begin{table}[!h]
\vspace{-0.5cm}
\caption{\small{Ablation studies on the Custom101 dataset.}}
\label{table:ablation}
\centering
\small
\setlength{\tabcolsep}{4pt}
\renewcommand{\arraystretch}{1.1}
\scalebox{0.7}{
\begin{tabular}{lcccccc}
\hline
\textbf{Metric} 
& \multicolumn{2}{c}{\textbf{RL}}
& \multicolumn{2}{c}{\textbf{Loss}}
& \multicolumn{2}{c}{\textbf{Finetune}} \\
\cline{2-7}
& w/o RL & w/ RL
& $L_{\text{MSE}}$
& $L_{\text{MSE}}{+}L_{\text{CE}}$
& Whole
& Cross-attn \\
\hline
Text align. (CLIP-T)  
& 0.12 & \textbf{0.59}
& 0.32 & \textbf{0.59}
& 0.33 & \textbf{0.59} \\
Image align. (CLIP-I)
& 0.20 & \textbf{0.63}
& 0.44 & \textbf{0.63}
& 0.42 & \textbf{0.63} \\
\hline
\end{tabular}}
\vspace{-0.5cm}
\end{table}

%% file: Section/5_Conclusion.tex
\vspace{0.5cm}
\section{Conclusion}
This work introduces a novel approach aimed at efficiently personalizing text-to-image diffusion models. Our method capitalizes on a Concept-Extraction network to create equivalents of the concept or subject depicted in a provided image within the textual-embedding space. By employing the predicted embedding, we can utilize text-to-image diffusion models to generate numerous variations of the specified subject. This strategy yields a marked reduction in inference time compared to methodologies reliant on test-time optimization. Through our experimentation, we have illustrated the capability of our method to generate high-quality and diverse images that faithfully represent the given subject, thereby enhancing zero-shot editability to a considerable extent. Furthermore, our approach effectively preserves subject details and uniqueness, positioning it as a viable component for future works in zero-shot customization.

%% file: Section/6_Supple.tex
\section{Supplementary Material}


In this supplementary material, we will provide the following details. 
\begin{enumerate}
    \item Training details.
    \item Computation complexity and timing analysis.
    \item Human Evaluation.
    \item Additional experimental results.
    \item Comparison with test-time optimization methods.
    \item Multi-attribute modification.
    \item Failure cases.
\end{enumerate}

\section{Training details}

\textbf{Train and Test split}

We start with providing the train-test split of our method. Since, this  method is generic and haven't explored much in our constrained setup (ID extractor for generic object and zero-shot customization in a single forward pass), we propose our train-test split to investigate into the problem. To this end, we split the custom101 dataset into train and test split as provided below. We train on identifiers on the train set and test on both the custom101 test split as well as totally out-of-distribution dreambooth dataset.

The training details of the proposed method are given here. Our method consist of two stages of training. 

Custom101\_Trainset  = [`actionfigure\_1', `actionfigure\_2', \\
`decoritems\_houseplant1', `decoritems\_houseplant2', `decoritems\_houseplant3', \\
`dish\_1', `flower\_1', `furniture\_chair1', `furniture\_chair2', `furniture\_chair3', \\
`furniture\_sofa1', `instrument\_1', `instrument\_music1', `instrument\_music2', \\
`jewelry\_earring', `jewelry\_ring', `luggage\_backpack1', `luggage\_purse1', \\
`luggage\_purse3', `luggage\_purse4', `person\_1', `person\_2', `pet\_cat1', \\
`pet\_cat2', `pet\_cat3', `pet\_cat7', `pet\_dog1', `pet\_dog2',`pet\_dog3', \\
`plushie\_2', `plushie\_bunny', `plushie\_cow', `plushie\_lobster', `plushie\_pink', \\`plushie\_sadangry', `plushie\_unicorn', `scene\_barn', `scene\_canal',\\
`scene\_castle', `scene\_lighthouse', `scene\_sculpture1',`scene\_waterfall', \\
`things\_book', `things\_book2',`things\_cup1', `things\_cup3', `things\_headphone1',\\
`things\_helmet', `things\_keychain1', `toy\_bear', `toy\_pikachu1', `toy\_unicorn', \\`transport\_bike', `transport\_car1', `transport\_car2', `transport\_car3',\\
`transport\_car4', `transport\_car5', `transport\_car6', `transport\_car7', \\
`transport\_car10', `transport\_car11',`transport\_motorbike1', `transport\_tank',\\
`wearable\_glasses', `wearable\_jacket1', `wearable\_jacket2', `wearable\_shoes1', \\ `wearable\_shoes2', `wearable\_sunglasses1',`wearable\_sunglasses2']

Custom101\_Testset = [`actionfigure\_3', `decoritems\_lamp1', `decoritems\_vase2', \\`decoritems\_woodenpot', `dish\_2', `flower\_2', `furniture\_sofa2', \\
`furniture\_table1',`instrument\_music3', `luggage\_purse2', `person\_3',\\
`pet\_cat4', `pet\_cat5', `pet\_cat6', `pet\_dog4', `plushie\_dice', `plushie\_happysad', \\
`plushie\_panda', `plushie\_penguin', `plushie\_teddybear', `plushie\_tortoise',\\
`scene\_garden', `things\_bottle1', `things\_corkscrew', `things\_cup2', \\
`things\_headphone2', `toy\_gnome', `toy\_tablechair', `transport\_car8', `transport\_car9']

Next, we provide details of the training pipeline. For the text prompt templates, we used a random subset of the standard CLIP-Imagenet templates as shown below. The hyperparameters for each stage of training and inference are provided in Table.~\ref{tab:zsp_hparams}.

training prompt templates = [
    `a photo of a \{\}',
    `a rendering of a \{\}',
    `a cropped photo of the \{\}',
    `the photo of a \{\}',
    `a photo of a clean \{\}',
    `a photo of a dirty \{\}',
    `a dark photo of the \{\}',
    `a photo of my \{\}',
    `a photo of the cool \{\}',
    `a close-up photo of a \{\}',
    `a bright photo of the \{\}',
    `a cropped photo of a \{\}',
    `a photo of the \{\}',
    `a good photo of the \{\}',
    `a photo of one \{\}',
    `a close-up photo of the \{\}',
    `a rendition of the \{\}',
    `a photo of the clean \{\}',
    `a rendition of a \{\}',
    `a photo of a nice \{\}',
    `a good photo of a \{\}',
    `a photo of the nice \{\}',
    `a photo of the small \{\}',
    `a photo of the weird \{\}',
    `a photo of the large \{\}',
    `a photo of a cool \{\}',
    `a photo of a small \{\}', ....
]

\begin{table}[!h]
    \centering
    \caption{Hyperparameters}
    \scalebox{0.8}{
    \begin{tabular}{cc}
    \toprule
    \hline
    Hyperparameter & Values \\
    \hline
    Stage 1 : Textual inversion generation \\
    \midrule
    Size of Textual inversion embedding & 1280 \\
    Image and text feature fusion dim. & CLIP-I (512) + CLIP-T (512) \\
    \midrule
    Stage 2 : Training MLP \\
    \midrule
    CLIP Model & ``openai/clip-vit-base-patch32'' \\
    input dim. of MLP & 1024 \\
    hidden dim. of MLP & 1000 \\
    output dim. of MLP & 1280  \\
    Coefficient of $L_{CE}$ & 1 \\
    Coefficient of $L_{MSE}$ & 1 \\
    Learning rate & $1e^{-3}$\\
    Batch size (training MLP) & 64 \\
    \midrule
    Stage 3 : Finetune the diffusion model \\
    \midrule
    Learning rate & $1e^{-5}$\\
    Batch size (LDM) & 125\\
    Text-to-image diffusion model & LDM~\cite{rombach2022high} \\
    \midrule
    Inference \\
    \midrule
    DDIM steps & 50 \\
    Scale (DDIM) & 10.0 \\
    \bottomrule
    \hline
    \end{tabular}}
    \label{tab:zsp_hparams}
\end{table}

\section{Computation complexity and timing analysis}

Our primary goal is to customize images based on text-prompts in a single forward pass. Therefore, our method is super-fast and comparison of inference time is provided in Tab.~\ref{table:zsp_compare_time}.
Note that the experiments have been performed on a single A5000 machine with 24GB GPU.

\begin{table}[!h]
\small
\centering
\vspace{-0.2cm}
\caption{Comparison of Inference speed}
\scalebox{0.95}{
\begin{tabular}{lccccc}
\toprule
\hline
\textbf{Method} & {Type}  & {speed(s)} \\
\hline
{Text. Inv.~\cite{gal2022image}} & {Finetune} & {2400}\\
{DB~\cite{ruiz2022dreambooth}} & {Finetune}  & {1284}\\
{CD~\cite{kumari2022multi}} & {Finetune} & {400}\\
{Re-Imagen~\cite{chen2022re}} & {Zero-shot} & {10} \\
{ELITE~\cite{wei2023elite}} & {Zero-shot} & {4} \\
{BLIP-Diff~\cite{li2024blip}} & {Zero-shot} & {8} \\
{Ours} & {Zero-shot} & {2} \\
\bottomrule
\hline
\end{tabular}}
\label{table:zsp_compare_time}
\end{table}

\begin{figure}
    \centering
    \includegraphics[scale=0.6]{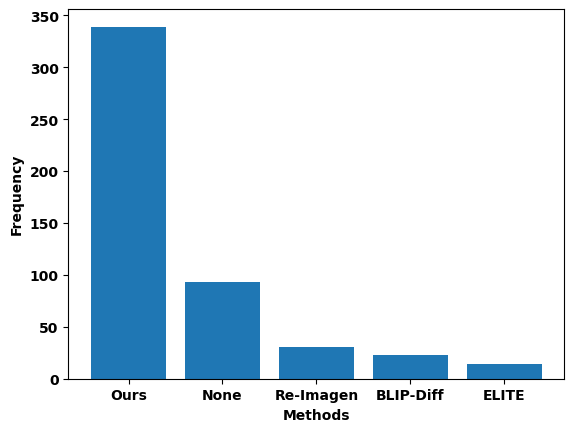}
    \caption{\small{Human evaluation comparison w.r.t test-time optimization methods}} 
    \label{fig:zsp_human_eval}
\end{figure}

\section{Human Evaluation}

Since the perceptual quality is quite subjective, automatic metrics do not correlate well with the perceptual studies~\cite{zhang2018unreasonable}. To verify that the improved scores actually correspond to better quality images,
we evaluate our approach using human preference study through Amazon Mechanical Turk.
We conduct human evaluation to compare our zero-shot method with SOTA zero-shot methods, viz., Re-Imagen~\cite{chen2022re}, ELITE~\cite{wei2023elite}, BLIP-Diff.~\cite{li2024blip} and the result is shown in Fig.~\ref{fig:zsp_human_eval}.
We provide randomly 20 examples of with random prompts from the Custom101 test dataset and asked the following question to amazon mechanical turks: ``which of the generated images is of best visual quality considering factors include image quality and preserving the identity of the original image?'' 
We evaluate this by 50 users, totalling 1000 questionnaires.
The available options are \{ `Re-Imagen', `ELITE', `BLIP-Diff', `None is satisfactory' \}.
The aggregate response shows that the generated images from our method are of better image quality compared to the baselines, as shown in Fig.~\ref{fig:zsp_human_eval}, where the average count of each method for all the observers are reported. By frequency, we mean how many times a method has chosen to be the best compared to others, and if none of the methods are satisfactory, then 'None is satisfactory' is chosen. We observe that our method significantly outperforms SOTA methods in human evaluation. 



\begin{figure}
    \centering
    \includegraphics[scale=0.33]{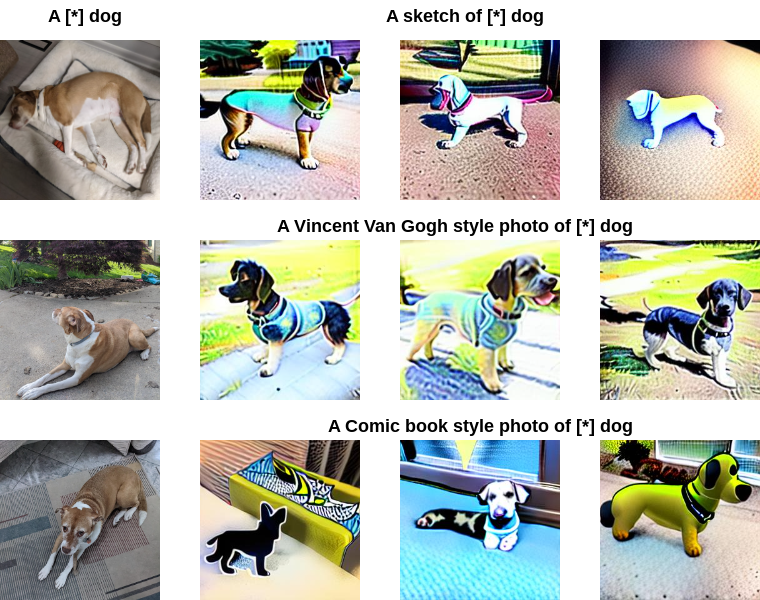}
    \caption{\small{Style transfer using our method}} 
    \label{fig:style_transfer}
\end{figure}


\begin{figure}
    \centering
    \includegraphics[scale=0.33]{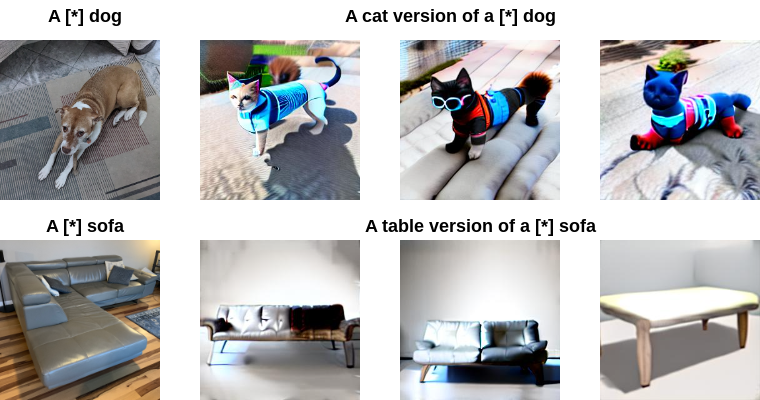}
    \caption{\small{Objectness transfer using our method}} 
    \label{fig:subject_transfer}
\end{figure}

\section{Additional experimental results}

\subsection{Subject modifications}
Using our fast zero-shot method, we can modify the subject to another subject based on prompt injection. E.g., in Fig.~\ref{fig:subject_transfer}, we start with a specific dog image, and generate a cat version of the dog by modifying the prompt as ``A cat version of a [*] dog''. We observe the generated images resembles both with the dog and also a cat. Therefore, our method can be used of complex object level editing, where editing requires knowledge of object level semantics. Similar trend can be followed when generating a ``table version of a sofa'' as shown in Fig.~\ref{fig:subject_transfer}

\subsection{Stylizing}
We can generate images of different styles as shown in Fig.~\ref{fig:style_transfer}, guided by the text prompt. In this instance, we are generating images of the [*] dog in sketch, Van Gogh and comic book styles. Thus, the method can be used for style transfer as well. 

\section{Comparison with test-time optimization methods}

We have qualitatively compare with multiple test-time optimization methods, viz., Re-Imagen~\cite{chen2022re}, ELITE~\cite{wei2023elite}, BLIP-Diff.~\cite{li2024blip} on the custom101 test set. The quantitative results are provided in the main paper. We have also performed the human evaluation and corresponding results in Fig.~\ref{fig:zsp_human_eval} clearly shows that our method improves upon the SOTA methods.  

\section{Multi-attribute modification}

\begin{figure}
    \centering
    \includegraphics[scale=0.33]{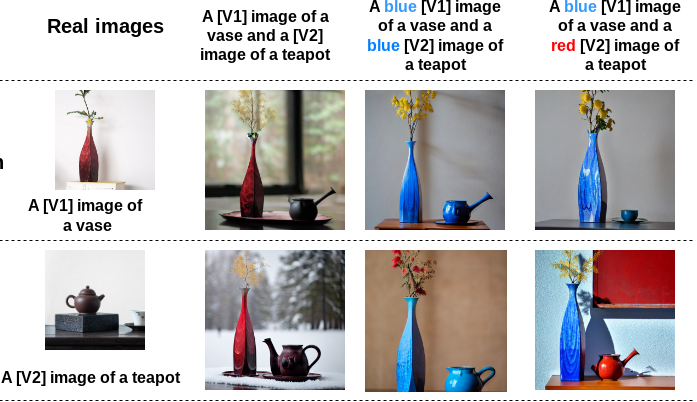}
    \caption{\small{Multiple attribute modification}} 
    \label{fig:multi_attr_mod}
\end{figure}

Our method is flexible to modify multiple attributes since we extract the concepts (textual inverison) in a single shot network. We have shown the multi-attribute modification in Fig.~\ref{fig:multi_attr_mod}. Hence, our method is more versatile and generic.

\section{Failure cases}

\begin{figure}
    \centering
    \includegraphics[scale=0.2]{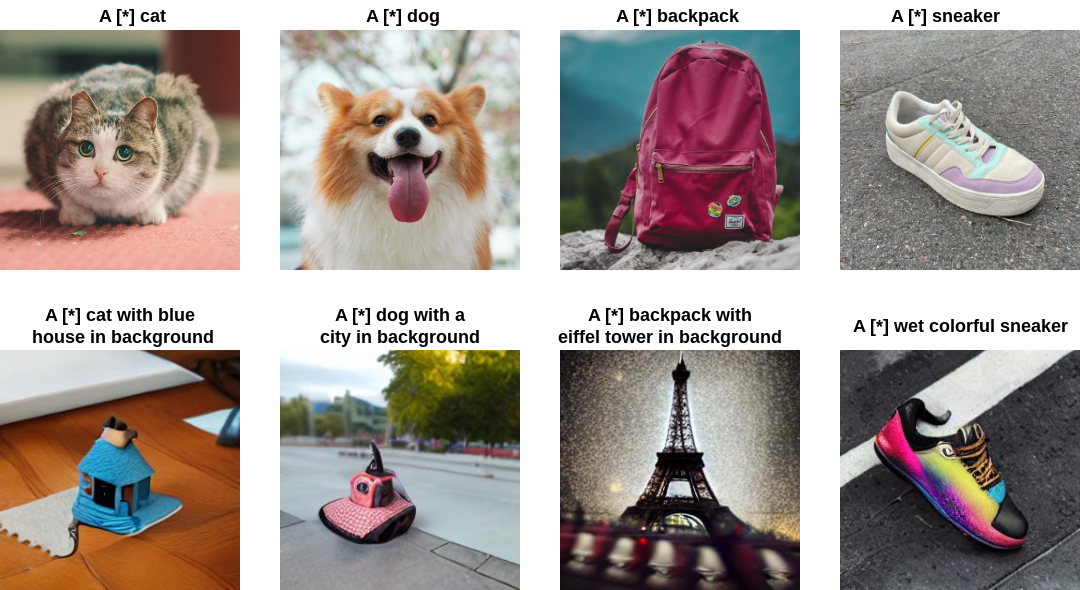}
    \caption{\small{Failure cases}} 
    \label{fig:zsp_supple_failure_cases}
\end{figure}

Some of the failure cases of our method have been shown in Fig.~\ref{fig:zsp_supple_failure_cases}. We observe that our method sometimes either misses the object (in case of the cat / backpack image modification), modifies the object (dog image modification) or there could be prompt inconsistencies (modification of sneaker). 
Some potential reasons could be the less diversity captured by the MLP and limited number of textual inversions used for training the MLP, which makes the problem even more challenging.
We would like to mitigate this issues in our future works.

